\documentclass[10pt,twocolumn,letterpaper]{article}

\usepackage[pagenumbers]{cvpr} 

\usepackage{graphicx}
\usepackage{amsmath}
\usepackage{amssymb}
\usepackage{booktabs}

\usepackage{amsthm}
\usepackage{multirow}

\usepackage[pagebackref,breaklinks,colorlinks]{hyperref}

\usepackage[accsupp]{axessibility}  

\usepackage[capitalize]{cleveref}
\crefname{section}{Sec.}{Secs.}
\Crefname{section}{Section}{Sections}
\Crefname{table}{Table}{Tables}
\crefname{table}{Tab.}{Tabs.}

\def\cvprPaperID{5645} 
\def\confName{CVPR}
\def\confYear{2023}

\begin{document}

\title{Adaptive Spot-Guided Transformer for Consistent Local Feature Matching}

\author{Jiahuan Yu\footnotemark[1], Jiahao Chang\footnotemark[1], Jianfeng He, Tianzhu Zhang\footnotemark[2], Feng Wu\\
{University of Science and Technology of China}
}




\twocolumn[{%
    \renewcommand\twocolumn[1][]{#1}%
	\maketitle
	\begin{center}
	\includegraphics[width=1.0\linewidth]{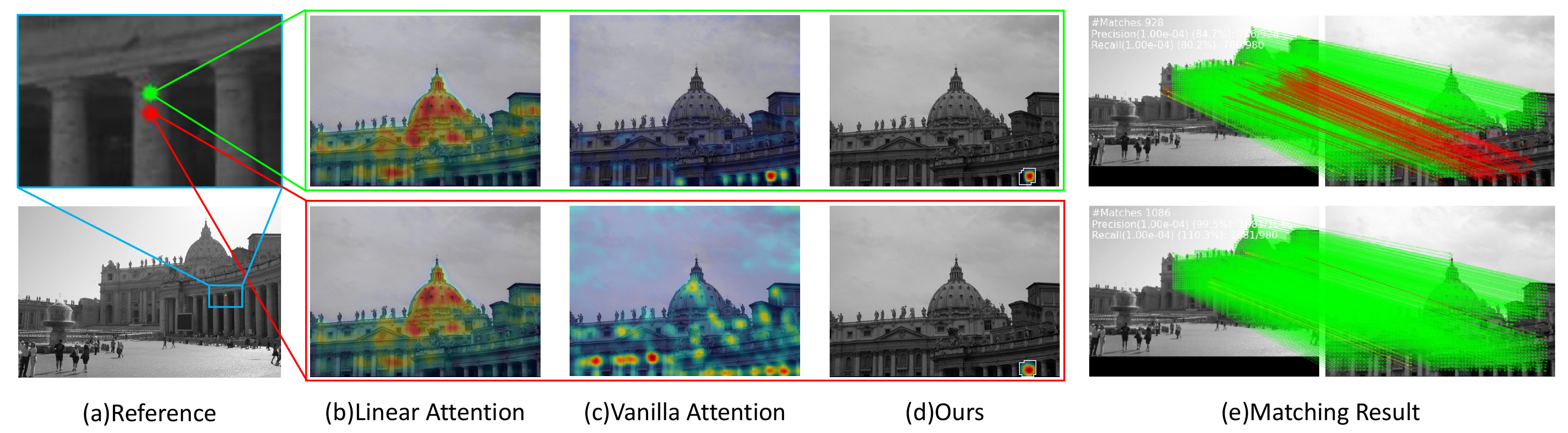}
	\captionof{figure}{The visualization of the cross attention heatmaps and matching results.
    We sample two similar adjacent points in the reference image (a), marked with green and red.
    (b) are two heatmaps of the linear cross attention in LoFTR~\cite{sun2021loftr} when green and red pixels are queries.
    (c) are two heatmaps obtained from the vanilla cross attention.
    (d) are two heatmaps generated by our spot-guided attention.
    (e) are the comparison of the final matching results produced by LoFTR~\cite{sun2021loftr} (top) and our method (down).}
	\label{fig:motivation}
	\end{center}
}]


\renewcommand{\thefootnote}{\fnsymbol{footnote}} 
\footnotetext[1]{Equal Contribution} 
\footnotetext[2]{Corresponding Author} 

\begin{abstract}
Local feature matching aims at finding correspondences between a pair of images.
Although current detector-free methods leverage Transformer architecture to obtain an impressive performance, few works consider maintaining local consistency.
Meanwhile, most methods struggle with large scale variations.
To deal with the above issues, we propose Adaptive Spot-Guided Transformer (ASTR) for local feature matching, which jointly models the local consistency and scale variations in a unified coarse-to-fine architecture.
The proposed ASTR enjoys several merits.
First, we design a spot-guided aggregation module to avoid interfering with irrelevant areas during feature aggregation.
Second, we design an adaptive scaling module to adjust the size of grids according to the calculated depth information at fine stage.
Extensive experimental results on five standard benchmarks demonstrate that our ASTR performs favorably against state-of-the-art methods.
Our code will be released on \url{https://astr2023.github.io}.

\end{abstract}

\section{Introduction}
\label{sec:intro}
Local feature matching (LFM) is a fundamental task in computer vision, which aims to establish correspondence for local features across image pairs.
As a basis for many 3D vision tasks, local feature matching can be applied in Structure-from-Motion (SfM)~\cite{schonberger2016structure}, 3D reconstruction~\cite{dai2017bundlefusion}, visual localization~\cite{sattler2018benchmarking, taira2018inloc}, and pose estimation~\cite{grabner20183d, persson2018lambda}.
%
%
Because of its broad applications, local feature matching has attracted substantial attention and facilitated the development of many researches~\cite{detone2018superpoint, li20dualrc, r2d2, rocco2018neighbourhood, sun2021loftr}.
%
However, finding consistent and accurate matches is
still difficult due to various challenging factors such as illumination variations, scale changes, poor textures, and repetitive patterns.

To deal with the above challenges, numerous matching methods have been proposed, which can be generally categorized into two major groups, including detector-based matching methods~\cite{barroso2019key, detone2018superpoint, dusmanu2019d2, ono2018lf, r2d2, sarlin2020superglue} and detector-free matching methods~\cite{huang2019dynamic, li20dualrc, rocco2020efficient, rocco2018neighbourhood, sun2021loftr, chen2022aspanformer}.
Detector-based matching methods require to first design a keypoint detector to extract the keypoints between two images, and then establish matches between these extracted keypoints.
%
The quality of detected keypoints will significantly  affect  the performance of detector-based matching methods.
Therefore, many works aim to improve keypoint detection through multi-scale detection~\cite{luo2020aslfeat}, repeatable and reliable verification~\cite{r2d2}.
%
Thanks to the high-quality keypoints detected, these methods can achieve satisfactory performance while maintaining high computational and memory efficiency.
%
%
However, these detector-based matching methods may have difficulty in finding reliable matches in textureless areas, where keypoints are challenging to detect.
Differently, detector-free matching methods do not need to detect keypoints and try to establish pixel-level matches between local features.
In this way, it is possible to establish matches in the texture-less areas.
Due to the power of attention in capturing long-distance dependencies, many Transformer-based methods~\cite{sun2021loftr, tang2022quadtree, wang2022matchformer, chen2022aspanformer} have emerged in recent years.
As a representative work,  considering the computation and memory costs, 
LoFTR~\cite{sun2021loftr} applies Linear Transformer~\cite{katharopoulos2020transformers} to aggregate global features at the coarse stage and then crops fixed-size grids for further refinement.
To alleviate the problem caused by scale changes, COTR~\cite{jiang2021cotr} calculate the co-visible area iteratively through attention mechanism.
The promising performance of Transformer-based methods proves that attention mechanism is effective on local feature matching.
Nevertheless, some recent works~\cite{li2022depthformer, yang2021transformer} indicate Transformer lacks spatial inductive bias for continuous dense prediction tasks, which may cause inconsistent local matching results. 


By studying the previous matching methods, we sum up two
issues that are imperative for obtaining the dense correspondence between images.
(1) \textbf{How to maintain local consistency.}
The correct matching result usually satisfies the local matching consistency, i.e., for two similar adjacent pixels, their matching points are also extremely close to each other.
Existing methods~\cite{sun2021loftr, wang2022matchformer,jiang2021cotr} utilize global attention in feature aggregation, introducing many irrelevant regions that affect feature updates.
Some pixels are disturbed by noisy or similar areas and aggregate information from wrong regions, leading to false matching results.
As shown in Figure~\ref{fig:motivation} (b), for two adjacent similar pixels, highlighted regions of global linear attention are decentralized and inconsistent with each other.
The inconsistency is also present in vanilla attention (see Figure~\ref{fig:motivation} (c)).
Therefore, it is necessary to utilize local consistency to focus the attention area on the correct place. 
(2) \textbf{How to handle scale variation.}
In a coarse-to-fine architecture, since the attention mechanism at the coarse stage is not sensitive to scale variations, we should focus on the fine stage.
Previous methods~\cite{li20dualrc, sun2021loftr, wang2022matchformer, chen2022aspanformer} select fixed-size grids for matching at the fine stage.
However, when the scale varies too much across images, the correct match point may be out of the range of the grid, resulting in matching failure.
Hence, the scheme of cropping grids should be adaptively adjusted according to scale variation across views.

To deal with the above issues, we propose a novel Adaptive Spot-guided Transformer (ASTR) for consistent local feature matching, including a spot-guided aggregation module and an adaptive scaling module.
In the \textbf{spot-guided aggregation module}, towards the goal of maintaining local consistency, we design a novel attention mechanism called spot-guided attention: each point is guided by similar high-confidence points around it, focusing on a local candidate region at each layer.
Here, we also adopt global features to enhance the matching ability of the network in the candidate regions.
Specifically, for any point $p$, we pick the points with high feature similarity and matching confidence in the local area.
Their corresponding matching regions are used for the next attention of point $p$.
In addition, global features are applied to help the network to make judgments.
The coarse feature maps are iteratively updated in the above way.
With our spot-guided aggregation module, the red and green pixels are guided to the correct area, avoiding the interference of repetitive patterns (see Figure~\ref{fig:motivation} (d)).
In Figure~\ref{fig:motivation} (e), our ASTR produces more accurate matching results, which maintains local matching consistency. 
In the \textbf{adaptive scaling module}, to fully account of possible scale variations, we attempt to adaptively crop different sizes of grids for alignment.
In detail, we compute the corresponding depth map using the coarse matching result and leverage the depth information to crop adaptive size grids from the high-resolution feature maps for fine matching.

The contributions of our method could be summarized into three-fold:
(1) We propose a novel Adaptive Spot-guided Transformer (ASTR) for  local feature matching, including a spot-guided aggregation module and an adaptive scaling module.
(2) 
%
We design a spot-guided aggregation module that can maintain local consistency and be unaffected by irrelevant regions while aggregating features.
Our  adaptive scaling module is able to leverage depth information to adaptively crop different size grids for refinement.
(3) Extensive experimental results on five challenging benchmarks show that our proposed method performs favorably against state-of-the-art image matching methods.

\section{Related Work}
\label{sec:rw}
In this section, we briefly review several research lines that are related to sparse matching methods, dense matching methods, and vision Transformer.


\noindent\textbf{Local Feature Matching}.
Local feature matching can categorized into detector-based and detector-free methods.
Detector-based methods can be divided into three stages: feature detection, feature description, and feature matching.
SIFT~\cite{lowe2004distinctive} and ORB~\cite{rublee2011orb} are the most popular hand-crafted local features, while learning-based methods~\cite{r2d2,dusmanu2019d2,detone2018superpoint,rublee2011orb,barroso2019key,zhou2017progressive,he2021consistency} also obtain good performance improvement compared to classical methods.
There are also some works focusing on improving the feature matching stage.
D2Net~\cite{dusmanu2019d2} fuses the detection and description stages.
R2D2~\cite{r2d2} attempts to train a network to find reliable and repeatable local features.
SuperGlue~\cite{sarlin2020superglue} proposes an attention-based GNN network to update extracted local features in alternating self and cross attentions.
However, detector-based methods rely on local feature extractors, which limits the performance in challenging scenarios such as repetitive textures, weak textures, and illumination changes.
Unlike detector-based approaches, detector-free approaches do not require a local feature detector, but find dense feature matching between pixels directly.
The classical methods~\cite{lucas1981iterative,horn1981determining} exists, but few of them outperform detector-based methods.
Learning-based methods change the game, which can be divided into cost-volume-based methods~\cite{rocco2018neighbourhood,li20dualrc,truong2020glu,truong2023pdc} and Transformer-based methods~\cite{jiang2021cotr,sun2021loftr,wang2022matchformer,chen2022aspanformer,huang2022adaptive,chen2022guide}.
Good performance have been achieved by cost-volume-based methods, but most of them are limited by the small receptive field of CNN, which is overcome by Transformer-based methods~\cite{sun2021loftr}.
Detector-free methods attain better performance in local feature matching, so we adopt this paradigm as the baseline.

\noindent\textbf{Vision Transformer.}
Transformer~\cite{vaswani2017attention} has been proven to be better at capturing long-range correlations than CNN in vision tasks~\cite{meng2022adversarial,carion2020end,meng2022task}.
Despite the great success, the computational cost of vanilla attention at high resolution is unacceptable, so some approximations~\cite{katharopoulos2020transformers,liu2021swin,tang2022quadtree,wang2020linformer} have been proposed, which inevitably leads to performance degradation.
Linear Attention~\cite{katharopoulos2020transformers} approximates softmax with ELU~\cite{clevert2015fast} to reduce the computational complexity to linear but degrades the focusing ability of attention. Swin-Transformer~\cite{liu2021swin} limits attention in local windows, which harms the ability to establish long-range associations.
At the same time, QuadTree~\cite{tang2022quadtree} calculates attention in a coarse-to-fine manner, and ASpanFormer~\cite{chen2022aspanformer} proposes an adaptive method for selecting attention spans, but few of them consider local consistency.
Different from the existing attention mechanism, we explicitly model local consistency in our spot-guided attention without introducing excessive computation and memory costs.

\noindent\textbf{Local Feature Matching with Scale Invariance.}
Scale variation is one of the main challenges faced by local feature matching.
Many works have explored solutions.
Hand-crafted local features~\cite{rublee2011orb,liu2010sift,bay2008speeded,rosten2006machine} use Gaussian pyramid model to alleviate the problem.
Following the hand-crafted methods, Some learning-based descriptors~\cite{r2d2,barroso2019key,barroso2020hdd,luo2020aslfeat,liu2021densernet,zhou2017progressive} also use the multi-scale representation.
ScaleNet~\cite{barroso2022scalenet} and Scale-Net~\cite{fu2021scale}, instead, try to directly estimate the scale ratio.
Another popular paradigm is to perform a wrap or scaling operation to eliminate the distortion caused by the scale variance.
GeoWrap~\cite{berton2021viewpoint} introduces a homography regression and warps images to increase overlap area.
OETR~\cite{chen2022guide} limits the keypoint detection in estimated overlap areas.
COTR~\cite{jiang2021cotr} estimates scale by finding co-visible regions, and then finds correspondence by recursively zooming.
However, most of above methods require significant modifications to the network architecture, and introduce additional computation overhead.
Therefore, we design a fully pluggable, lightweight and training-free module for coarse-to-fine architecture.

\section{Our Approach}
\label{sec:method}
\begin{figure*}[t]
    \centering
    \includegraphics[width=0.95\linewidth]{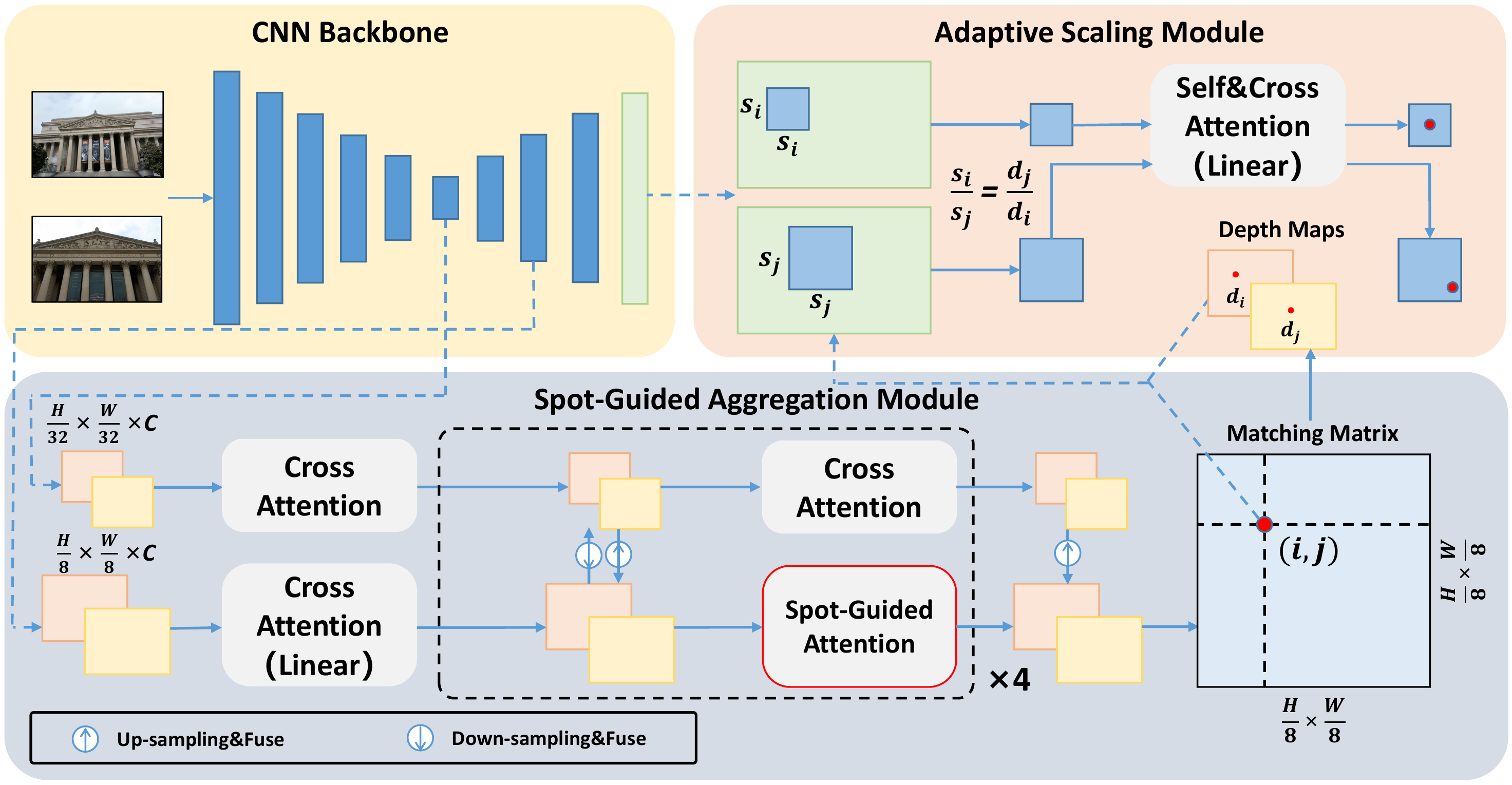}
    \caption{
    The architecture of ASTR.
    Our ASTR consists of two major components: spot-guided aggregation module and the adaptive scaling module.
    ``Cross Attention'' means vanilla cross attention, unless otherwise stated.
    Please refer to the text for detailed architecture.
    }\label{fig:method}
\end{figure*}
In this section, we present our proposed Adaptive Spot-guided Transformer (ASTR) for Consistent Local Feature Matching.
The overall architecture is illustrated in Figure~\ref{fig:method}.
\subsection{Overview}\label{3.1}
As shown in Figure~\ref{fig:method}, the proposed ASTR mainly consists of two modules, including a spot-guided aggregation module and an adaptive scaling module.
Here we give a brief introduction to the entire process.
Given an image pair $I_{Ref}$ and $I_{Src}$, to start with, we extract multi-scale feature maps of each image through a shared Feature Pyramid Network (FPN)~\cite{lin2017feature}.
We denote feature maps with the size of $1/i$ as $F^{1/i} = \{F^{1/i}_{ref}, F^{1/i}_{src}\}$.
Then, $F^{1/32}$ and $F^{1/8}$ are fed into the spot-guided aggregation module for coarse-matching and depth maps.
Here, the coarse matching result is acquired in three phases.
First, we need to compute the similarity matrix, which can be given by $S(i,j) = \tau \langle F^{1/8}_{ref}(i), F^{1/8}_{src}(j) \rangle$ with flattened features, where $\tau$ is the temperature coefficient.
Then we perform dural-softmax operator on $S$ to calculate matching matrix $\mathrm{P_c}$:
\begin{small}
\begin{equation}\label{eq:epc7}
    \noindent \mathrm{P_c}(i, j) = \mathrm{softmax}(S(i, :))(i, j) \cdot \mathrm{softmax}(S(:, j))(i, j).
\end{equation}
\end{small}
Finally, we use the mutual nearest neighbor strategy and the threshold $\theta_c$ to filter out the coarse-matching result $M_c$.
According to depth information and coarse-matching result, we can crop different size grids on the high-resolution feature map $F^{1/2}$.
After linear self and cross attention layers, features of the cropped grids are used to produce the final fine-level matching result.
\subsection{Spot-Guided Aggregation Module}\label{3.2}
Correct matching always satisfies the local matching consistency, i.e., the matching points of two similar adjacent pixels are also close to each other in the other image.
When humans establish dense matches between two images, they will first scan through the two images quickly and keep in mind some landmarks that are easier to match correctly.
For those trouble points similar to surrounding landmarks, it is not easy to obtain correct matches in the beginning.
But now, they can focus attention around the matching points of landmarks to revisit trouble points' matches.
In this way, more correctly matched landmarks are obtained.
After several iterations of the above process, eventually, they will get the matching result for the whole image.
Inspired by this idea, we design  a spot-guided aggregation module.
Section~\ref{3.2.1} introduces the preliminaries of vanilla attention and linear attention.
Section~\ref{3.2.2} describes our spot-guided attention mechanism.
Section~\ref{3.2.3} demonstrates the design of the entire spot-guided aggregation module.

\subsubsection{Preliminaries}\label{3.2.1}
The calculation of vanilla attention requires three inputs: query $Q$, key $K$, and value $V$.
The output of vanilla attention is a weighted sum of the value, where the weight matrix is determined by the query and its corresponding key.
The process can be described as
\begin{small}
\begin{equation}\label{eq:epc1}
    \mathrm{Attention}(Q, K, V) = \mathrm{softmax}(QK^T)V.
\end{equation}
\end{small}
However, in vision tasks, the size of the weight matrix $\mathrm{softmax}(QK^T)$ increases quadratically as the image resolution grows.
When the image resolution is large, the memory and computational cost of vanilla attention is unacceptable.
To solve this problem, 
Linear attention ~\cite{katharopoulos2020transformers} is proposed to replace the softmax operator with the product of two kernel functions:
\begin{small}
\begin{equation}\label{eq:epc2}
    \mathrm{Linear\_attention}(Q, K, V) = \phi (Q) (\phi (K^T)V),
\end{equation}
\end{small}
where $\phi(\cdot) = \mathrm{elu}(\cdot) + 1$.
Since the number of feature channels is much smaller than the number of pixels, the computational complexity is reduced from quadratic to linear.

\begin{figure}[t]
    \centering
    \includegraphics[width=1.0\linewidth]{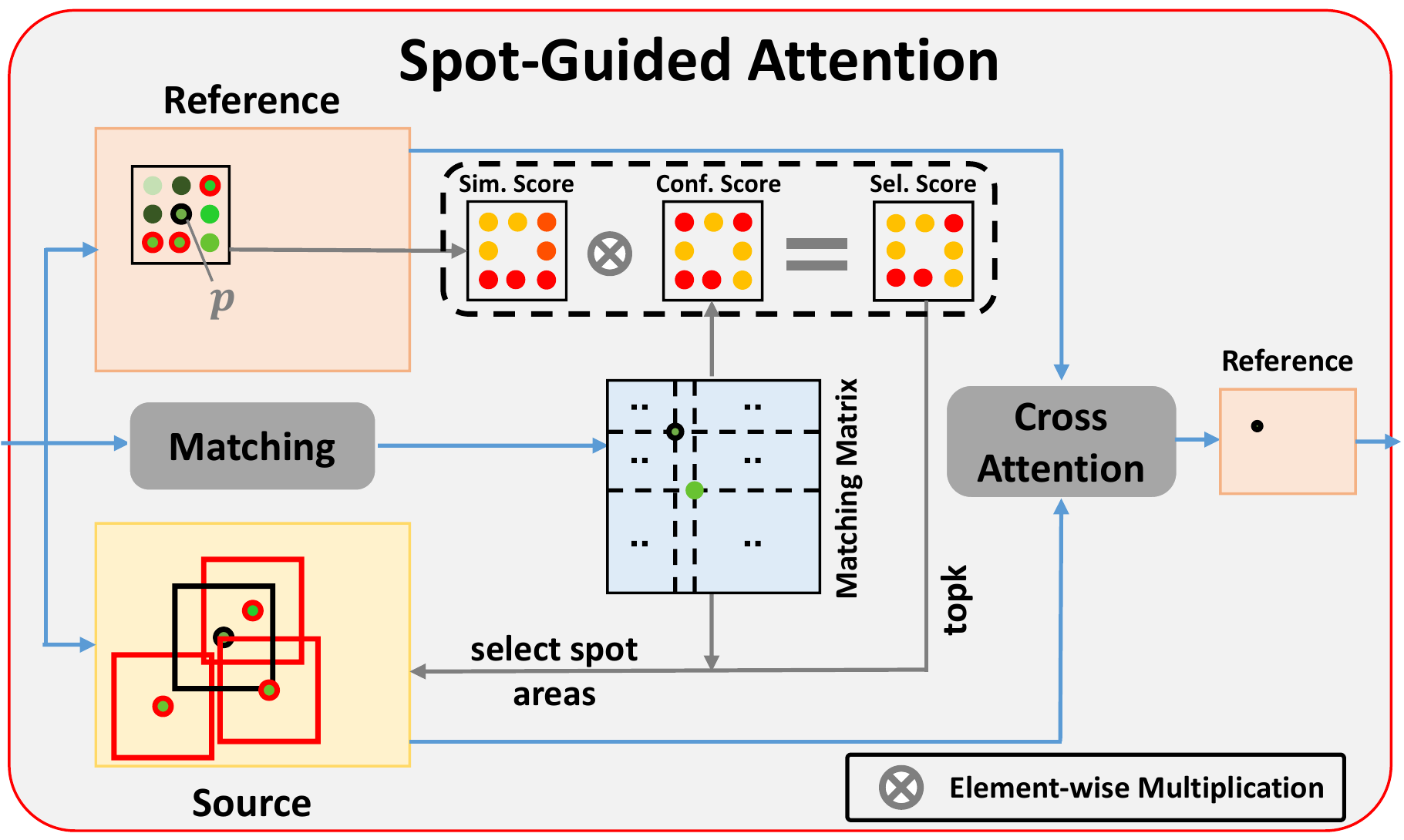}
    \caption{
    The illustration of our spot-guided attention.
    }\label{fig:spot}
\end{figure}

\subsubsection{Spot-Guided Attention}\label{3.2.2}
It is known from the local matching consistency that the matching points of similar adjacent pixels are also close to each other.
In Figure~\ref{fig:spot}, we illustrate the case that the reference image as query aggregates features from the source image.
Given reference and source feature maps $F^{1/8} = \{F^{1/8}_{ref}, F^{1/8}_{src}\}$, 
we compute a matching matrix $P_s$ across images.
For any pixel $p$ in Figure~\ref{fig:spot}, we first compute the similarity score $\mathrm{S_{sim}}(p) \in R^{l^2 - 1}$ between $p$ and other pixels in the $l \times l$ area around $p$.
Specifically, the similarity score can be obtained as
\begin{small}
\begin{equation}\label{eq:ep3}
\noindent \mathrm{S_{sim}}\left(p\right) = \mathop{\mathrm{softmax}}\limits_{i}\left(\left\{\langle F^{1/8}_{ref}(p), F^{1/8}_{src}(p_i) \rangle\right\}_{p_i \in N(p)}\right),
\end{equation}
\end{small}
where $\langle \cdot, \cdot \rangle$ is the inner product, and $N(p)$ is the set of pixels in the $l \times l$ field around pixel $p$. 
In addition, we should also consider the reliability of points in $N(p)$.
For each $p_i \in N(p)$, confidence can be viewed as the highest similarity to all pixels on the source images.
Meanwhile, we can also get the matching point position of $p_i$, denoted as $\mathrm{Loc}(p_i)$.
Hence, $\mathrm{Loc}(p_i)$ and confidence score $\mathrm{S_{conf}}(p) \in R^{l^2 - 1}$ can be computed in the following way:
\begin{small}
\begin{equation}\label{eq:epc4}
    \begin{aligned}
    &\mathrm{S_{conf}}\left(p\right) = \left\{\mathop{\mathrm{max}}\left(\mathrm{P_s}\left(p_i,:\right)\right)\right\}_{p_i \in N(p)}.
    \\
    &\mathrm{Loc}\left(p_i\right) = \mathop{\mathrm{argmax}}\left(\mathrm{P_s}\left(p_i,:\right)\right),
    \end{aligned}
\end{equation}
\end{small}
Combining two scores, we select $p$ and top-k points $\mathrm{Topk}(p)$ whose matching points are used as seed points $\mathrm{Seed}(p)$:
\begin{small}
\begin{equation}\label{eq:epc5}
    \begin{aligned}
    &\mathrm{Topk}(p) = \{p\} \cup \mathrm{topk}\{\mathrm{S_{sim}}(p) \cdot \mathrm{S_{conf}}(p)\},
    \\
    &\mathrm{Seed}(p) = \{\mathrm{Loc}(q)\}_{q \in \mathrm{Topk}(p)},
    \end{aligned}
\end{equation}
\end{small}
Following that, we extend $l \times l$ regions centered on these seed points $\mathrm{Seed}(p)$ on $I_{src}$, which are the spot areas of $p$.
Finally, cross attention is performed between $p$ and corresponding spot areas.
After exchanging the source image and the reference image, the source feature map is updated in the same way. 

\subsubsection{Spot-Guided Feature Aggregation}\label{3.2.3}
For the input features $F^{1/32}$ and $F^{1/8}$, $F^{1/32}$ is updated by vanilla cross attention, and $F^{1/8}$ is updated by linear cross attention for initialization.
Then, two features of different resolutions are fed into the spot-guided aggregation blocks.
In each block, $F^{1/32}$ and $F^{1/8}$ are first fused into each other in the following way:
\begin{small}
\begin{equation}\label{eq:epc6}
    \begin{aligned}
    &\hat{F}^{1/32} = F^{1/32} + \mathrm{Conv_{1 \times 1}}(\mathrm{Down}(F^{1/8})),
    \\
    &\hat{F}^{1/8} = F^{1/8} + \mathrm{Conv_{1 \times 1}}(\mathrm{Up}(F^{1/32})),
    \end{aligned}
\end{equation}
\end{small}
where $\hat{F}^{1/32}$ and $\hat{F}^{1/8}$ are features after fusion. $\mathrm{Down}(\cdot)$ and $\mathrm{Up}(\cdot)$ are downsampling and upsampling.
And then, $\hat{F}^{1/32}$ aggregate features across images by vanilla attention.
In the meantime, $\hat{F}^{1/8}$ aggregate features across images by spot-guided attention.
After four spot-guided aggregation blocks, $1/32$-resolution features are fused into $1/8$-resolution features, which are used to obtain the coarse-matching result $M_c$.



\begin{figure}[t]
    \centering
    \includegraphics[width=0.8\linewidth]{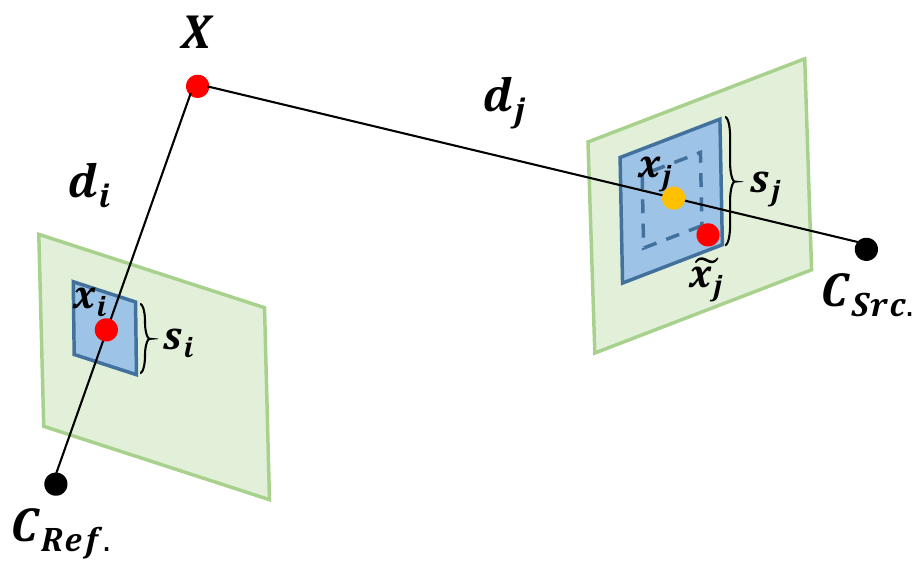}
    \caption{
    The illustration of our adaptive scaling module.
    On the left is the reference image, whose optical center is $C_{Ref}$.
    On the right is the source image, whose optical center is $C_{Src}$.
    $x_i$ and $x_j$ are the projections of the real-world point $X$.
    }\label{fig:adaptive_scaling}
\end{figure}

\subsection{Adaptive Scaling Module}\label{3.3}
At the fine stage, previous methods usually crop fixed-size grids based on the coarse matching result.
When there is a large scale variation, fine matching may fail since the ground-truth matching points are out of grids.
Thus, we refer to depth information to adaptively crop grids of different sizes between images.
Section~\ref{3.3.1} describes the way to obtain depth information from the coarse-matching result.
Section~\ref{3.3.2} demonstrates the process of adaptively cropping grids.

\subsubsection{Depth Information}\label{3.3.1}

With the coarse-level matching result, we can obtain the relative pose of two images $\{R, T\}$ through RANSAC~\cite{fischler1981random}.
It should be noted that the $T$ calculated here has a scale uncertainty, i.e., $T_{real} = \alpha T$, where $\alpha$ is the scale factor.
Given the image coordinates of any pair of matching points $\{x_i, x_j\}$ from coarse-level matching result, they satisfy the following equation:
\begin{small}
\begin{equation}\label{eq:epc8}
\noindent d_j K^{-1}_j (x_j, 1)^T = d_i R K^{-1}_i (x_i, 1)^T + \alpha T ,
\end{equation}
\end{small}
where $d_i$ and $d_j$ are the depth values of $x_i$ and $x_j$.
$K_i$ and $K_j$ are corresponding camera intrinsics.
We let $p_i = R K^{-1}_i (x_i, 1)^T$ and $p_j = K^{-1}_j (x_j, 1)^T$.
From Equation~\eqref{eq:epc8} it can be deduced that:
\begin{small}
\begin{equation}\label{eq:epc9}
\begin{aligned}
&d_j p_j = d_i p_i + \alpha T, \\
\Rightarrow &\left\{
    \begin{array}{lr}
    (d_j / \alpha) p_j \wedge p_i = 0 + T \wedge p_i, \\
    0 = (d_i / \alpha) p_i \wedge p_j + T \wedge p_j,
    \end{array}
\right.
\\
\Rightarrow &\left\{
    \begin{array}{lr}
    d_j / \alpha = \mathrm{mean}(\mathrm{div}(T \wedge p_i, p_j \wedge p_i)), \\
    d_i / \alpha = \mathrm{mean}(\mathrm{div}(- T \wedge p_j, p_i \wedge p_j)),
    \end{array}
\right.
\end{aligned}
\end{equation}
\end{small}
where $\wedge$ indicates outer product.
$\mathrm{div}(\cdot, \cdot)$ denotes element-wise division between two vectors.
$\mathrm{mean}(\cdot)$ is the scalar mean of each component of a vector.
In this way, we have obtained depth information of $x_i$ and $x_j$ with scale uncertainty.

\subsubsection{Adaptive Scaling Strategy}\label{3.3.2}
As shown in Figure~\ref{fig:adaptive_scaling}, $x_i$ and $x_j$ are a pair of matching points at the coarse stage.
$d_i$ and $d_j$ are depth values of $x_i$ and $x_j$.
To begin with, we crop a $s_i \times s_i$ region centered on $x_i$.
When the scale changes too much, the correct matching point $\widetilde{x_j}$ may be beyond the $s_i \times s_i$ region around $x_j$.
Because everything looks small in the distance and big on the contrary, the size of cropped grid $s_j$ should satisfy:
\begin{small}
\begin{equation}\label{eq:epc10}
\noindent \frac{s_j}{s_i} = \frac{d_i}{d_j} = (\frac{d_i}{\alpha})(\frac{d_j}{\alpha})^{-1},
\end{equation}
\end{small}
Following the above approach, we can crop different sizes of grids adaptively according to the scale variation.
After the same refinement as LoFTR~\cite{sun2021loftr}, we get the final matching position $\widetilde{x_j}$ of $x_i$.

\subsection{Loss Function}\label{3.3}
Our loss function mainly consists of three parts, spot matching loss, coarse matching loss, and fine matching loss.
Spot matching loss is the cross entropy loss to supervise the matching matrix during spot-guided attention:
\begin{equation}\label{eq:epc11}
\noindent L_s = - \frac{1}{\left\lvert M^{gt}_c \right\rvert} \sum_{(i,j) \in M^{gt}_c} \log \mathrm{P_s}(i,j),
\end{equation}
where $M^{gt}_c$ is the ground truth matches at coarse resolution.
Coarse matching loss is also the cross entropy loss to supervise the coarse matching matrix:
\begin{equation}\label{eq:epc12}
\noindent L_c = - \frac{1}{\left\lvert M^{gt}_c \right\rvert} \sum_{(i,j) \in M^{gt}_c} \log \mathrm{P_c}(i,j).
\end{equation}
Fine matching loss $L_f$ is a weighted $L_2$ loss same as LoFTR~\cite{sun2021loftr}.
Therefore, our total loss is:
\begin{equation}\label{eq:epc14}
\noindent L_{total} = L_s + L_c + L_f.
\end{equation}

\section{Experiments}
\label{sec:expr}
In this section, we evaluate our ASTR with extensive experiments.
First of all, we introduce implementation details, followed by experiments on five benchmarks and some visualizations.
Finally, we conduct a series of ablation studies to verify the effectiveness of each component.

\subsection{Implementation Details}\label{4.1}
We implement the proposed model in Pytorch~\cite{paszke2019pytorch}.
Our ASTR is trained on the MegaDepth dataset~\cite{li2018megadepth}.
In the training phase, we input images with the size of $832 \times 832$ for training.
The CNN extractor is a deepened ResNet-18~\cite{he2016deep} with features at $1/32$ resolution.
In spot-guided attention, we set the kernel size of local region $l$ to 5 and $k$ to 4 in $\mathrm{topk}$.
Threshold $\theta_c$ in coarse matching is chosen to 0.2.
At the fine stage, window size $s_i$ in the reference image is fixed to 5, and window size $s_j$ in the source image will be adaptively calculated according to the depth information.
In particular, $s_j/s_i$ is clamped into $[1, 3]$.
Our network is trained for 15 epochs with a batch size of 8 by Adam~\cite{kingma2014adam} optimizer.
The initial learning rate is $1 \times 10^{-3}$.
In order to establish spot-guided attention efficiently, we implement a highly optimized general sparse attention operator based on CUDA.
Please refer to the Supplementary Material for more details about the operator.
\begin{table}[ht]
	\centering
	\small
	\caption{Evaluation on HPatches~\cite{balntas2017hpatches} for homography estimation.}
	\label{tab:HPatches_result}
	\vspace{-3mm}
	\scalebox{0.75}{
		\begin{tabular}{c l c c c c}
			\hline
            \multirow{2}{*}{Category} &\multicolumn{1}{c}{\multirow{2}{*}{Method}} &\multicolumn{3}{c}{Homography est. AUC} &\multirow{2}{*}{matches} \\
			\cline{3-5}
                    & &{@3px}  &{@5px}  &{@10px} \\
			\hline
            \multirow{5}{*}{Detector-based} &D2Net~\cite{dusmanu2019d2}+NN &23.2 &35.9 &53.6 &0.2K \\

			                                &R2D2~\cite{r2d2}+NN &50.6 &63.9 &76.8 &0.5K \\

                                            &DISK~\cite{tyszkiewicz2020disk}+NN &52.3 &64.9 &78.9 &1.1K \\

                                            &SP~\cite{detone2018superpoint}+SuperGlue~\cite{sarlin2020superglue} &53.9 &68.3 &81.7 &0.6K \\

                                            &Patch2Pix~\cite{zhou2021patch2pix} &46.4 &59.2 &73.1 & 1.0k \\
            \hline
            \multirow{6}{*}{Detector-free} &Sparse-NCNet~\cite{rocco2020efficient} &48.9 &54.2 &67.1 &1.0K \\ 

                                            & COTR~\cite{jiang2021cotr} &41.9 &57.7 &74.0 &1.0K \\
                                            
                                            &DRC-Net~\cite{li20dualrc} &50.6 &56.2 &68.3 &1.0K \\
                                            
                                            &LoFTR~\cite{sun2021loftr} &65.9 &75.6 &84.6 &1.0K \\           

                                            &PDC-Net+~\cite{truong2023pdc} &66.7 & 76.8 & 85.8 & 1.0k \\

			                                &\textbf{ASTR(ours)}       &\bf 71.7 &\bf 80.3 &\bf 88.0 &1.0K   \\
			\hline
		\end{tabular}
	}
\end{table}

\subsection{Homography Estimation}\label{4.2}
\textbf{Dataset and Metric.}
HPatches~\cite{balntas2017hpatches} is a popular benchmark for image matching.
Following~\cite{dusmanu2019d2} , we choose 56 sequences under significant viewpoint changes and 52 sequences with large illumination variation to evaluate the performance of our ASTR trained on MegaDepth~\cite{li2018megadepth}.
We use the same evaluation protocol as LoFTR~\cite{sun2021loftr}.
We report the area under the cumulative curve (AUC) of the corner error distance up to 3, 5, and 10 pixels, respectively.
We limit the maximum number of output matches to 1k.

\textbf{Results.}
In Table~\ref{tab:HPatches_result}, we can see that our ASTR achieves new state-of-the-art performance on HPatches~\cite{balntas2017hpatches} under all error thresholds, which strongly proves the effectiveness of our method.
ASTR outperforms the best method before ({PDC-net+}~\cite{truong2023pdc}), achieving a large margin of $\textbf{4.4\%}$ under 3 pixels, $\textbf{3.5\%}$ under 5 pixels, and $\textbf{2.5\%}$ under 10 pixels.
Thanks to the proposed spot-guided aggregation module and adaptive scaling module, our method can yield more accurate matches under extreme viewpoint and illumination variations.

\subsection{Relative Pose Estimation}\label{4.3}
\textbf{Dataset and Metric.}
We use MegaDepth~\cite{li2018megadepth} and ScanNet~\cite{dai2017scannet} to demonstrate the performance of our ASTR in relative pose estimation.
MegaDepth~\cite{li2018megadepth} is a large-scale outdoor dataset that contains 1 million internet images of 196 different outdoor scenes.
Each scene is reconstructed by COLMAP~\cite{schonberger2016structure}. 
Depth maps as intermediate results can be converted to ground truth matches.
We sample the same 1500 pairs as ~\cite{sun2021loftr} for testing.
All test images are resized such that their longer dimensions are 1216.
ScanNet~\cite{dai2017scannet} is usually used to validate the performance of indoor pose estimation.
It is composed of monocular sequences with ground truth poses and depth maps.
Wide baselines and extensive textureless regions in image pairs make ScanNet~\cite{dai2017scannet} challenging.
For a fair comparison, we follow the same testing pairs and evaluation protocol as ~\cite{sun2021loftr}.
And all test images are resized to $640 \times 480$.
Note that we use our ASTR trained on MegaDepth~\cite{li2018megadepth} to evaluate its performance on ScanNet~\cite{dai2017scannet}.   
We report the AUC of the pose error at thresholds $(5^{\circ}, 10^{\circ}, 20^{\circ})$, where pose error is the maximum angular error in rotation and translation.
The angular error is computed between the ground truth pose and the predicted pose.

\begin{table}[t]
	\centering
	\small
	\caption{Evaluation on MegaDepth~\cite{li2018megadepth} for outdoor relative position estimation.}
	\label{tab:MegaDepth_result}
	\vspace{-3mm}
	\scalebox{0.75}{
		\begin{tabular}{c l c c c c}
			\hline
            \multirow{2}{*}{Category} &\multicolumn{1}{c}{\multirow{2}{*}{Method}} &\multicolumn{3}{c}{Pose estimation AUC} \\
			\cline{3-5}
                    & &{@$5^\circ$}  &{@$10^\circ$}  &{@$20^\circ$} \\
			\hline
            \multirow{2}{*}{Detector-based} &SP~\cite{detone2018superpoint}+SuperGlue~\cite{sarlin2020superglue} &42.2 &59.0 &73.6 \\
            
                                            &SP~\cite{detone2018superpoint}+SGMNet~\cite{chen2021learning} &40.5 &59.0 &73.6 \\
            \hline
            \multirow{7}{*}{Detector-free} &DRC-Net~\cite{li20dualrc} &27.0 &42.9 &58.3 \\
                                            
                                            &PDC-Net+(H)~\cite{truong2023pdc}  &43.1  &61.9  &76.1 \\
                                            
                                            &LoFTR~\cite{sun2021loftr} &52.8 &69.2 &81.2 \\                          
                                            
                                            &MatchFormer~\cite{wang2022matchformer} &53.3 &69.7 &81.8 \\
                                            
                                            &QuadTree~\cite{tang2022quadtree} & 54.6 & 70.5 & 82.2 \\
                                            
                                            &ASpanFormer~\cite{chen2022aspanformer} &55.3 &71.5 &83.1 \\

			                                &\textbf{ASTR(ours)}       &\textbf{58.4} & \textbf{73.1} &\ \textbf{83.8}  \\
			\hline
		\end{tabular}
	}
\end{table}

\begin{table}[t]
	\centering
	\small
	\caption{Evaluation on ScanNet~\cite{dai2017scannet} for indoor relative position estimation. * indicates models trained on MegaDepth~\cite{li2018megadepth}.}
	\label{tab:ScanNet_result}
	\vspace{-3mm}
	\scalebox{0.75}{
		\begin{tabular}{c l c c c c}
			\hline
            \multirow{2}{*}{Category} &\multicolumn{1}{c}{\multirow{2}{*}{Method}} &\multicolumn{3}{c}{Pose estimation AUC} \\
			\cline{3-5}
                    & &{@$5^\circ$}  &{@$10^\circ$}  &{@$20^\circ$} \\
			\hline
            \multirow{3}{*}{Detector-based} &D2-Net~\cite{dusmanu2019d2}+NN &5.3 &14.5 &28.0 \\

											&SP~\cite{detone2018superpoint}+OANet~\cite{zhang2019learning} &11.8 &26.9 &43.9 \\
											
											&SP~\cite{detone2018superpoint}+SuperGlue~\cite{sarlin2020superglue} &16.2 &33.8 &51.8 \\
            \hline
            \multirow{5}{*}{Detector-free}  &DRC-Net~\cite{li20dualrc}* &7.7 &17.9 &30.5 \\                         
            
                                            &MatchFormer~\cite{wang2022matchformer}* &15.8  &32.0   &48.0 \\

											&LoFTR-OT~\cite{sun2021loftr}* &16.9 &33.6 &50.6 \\
                                            
                                            &Quadtree~\cite{tang2022quadtree}* &19.0 &37.3  &53.5 \\

			                                &\textbf{ASTR(ours)*}       & \textbf{19.4} & \textbf{37.6} & \textbf{54.4}  \\
			\hline
		\end{tabular}
	}
\end{table}

\begin{figure}[t]
    \centering
    \includegraphics[width=1.0\linewidth]{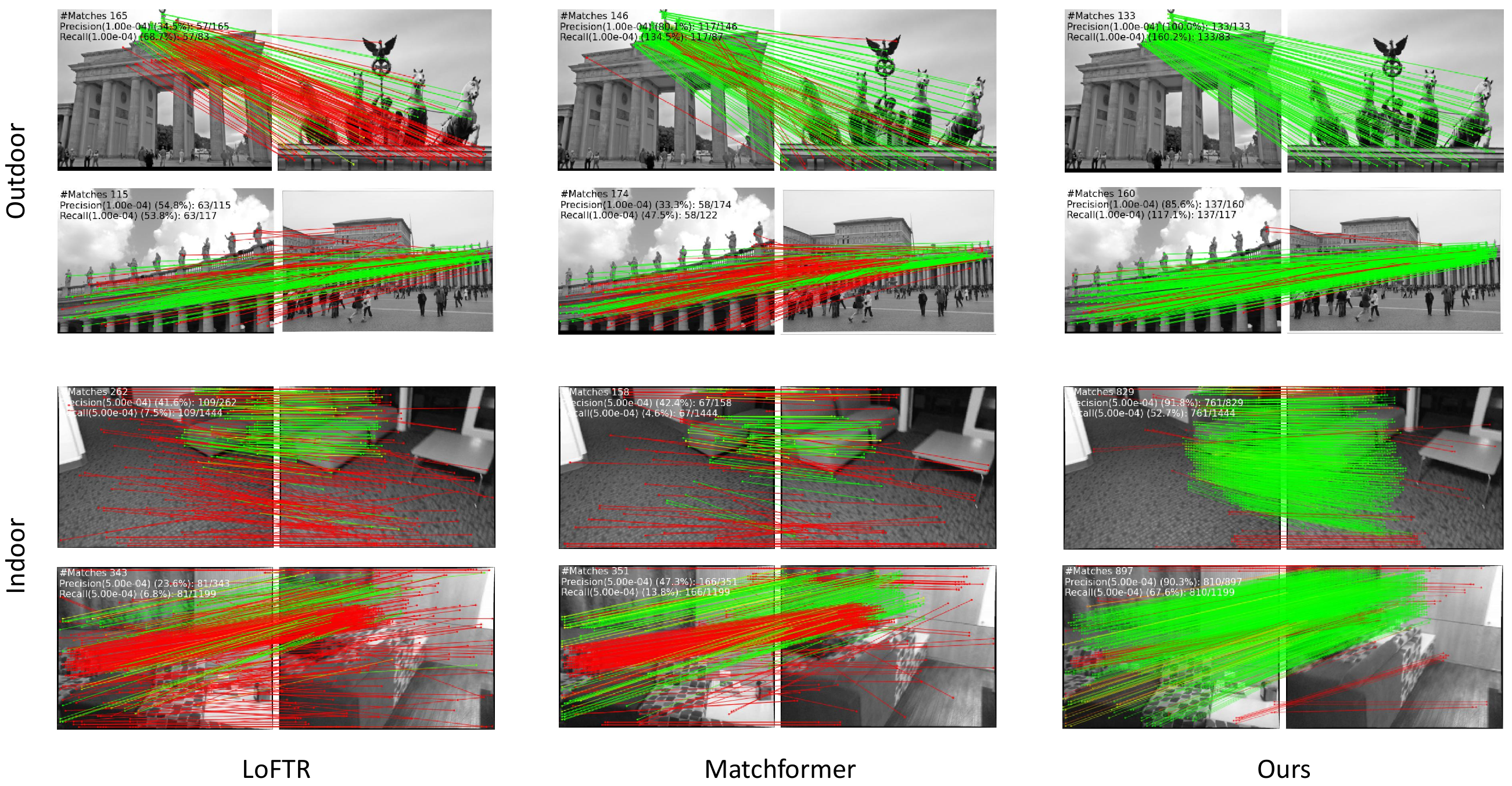}
    \vspace{-6mm}
	\caption{
	Qualitative results of dense matching on MegaDepth~\cite{li2018megadepth} and ScanNet~\cite{dai2017scannet}.
    }\label{fig:qualitative}
    \vspace{-3mm}
\end{figure}

\textbf{Results.}
As shown in Table~\ref{tab:MegaDepth_result}, our ASTR outperforms other state-of-the-art methods on MegaDepth~\cite{li2018megadepth}.
In particular, our ASTR improves by $\textbf{3.1\%}$ in AUC$@5^{\circ}$ and $\textbf{1.6\%}$ in AUC$@10^{\circ}$.
Table~\ref{tab:ScanNet_result} summarizes the performance comparison between the proposed ASTR and state-of-the-art methods on ScanNet~\cite{dai2017scannet}.
Our ASTR ranks first when only considering models not trained on ScanNet~\cite{dai2017scannet}, indicating the impressive generalization of our method.
Thanks to the proposed spot-guided aggregation module and adaptive scaling module, our method can yield more correct matches, resulting in more accurate pose estimation.
In order to further demonstrate the effectiveness of our ASTR, in Figure~\ref{fig:qualitative}, we visually demonstrate the comparison with other methods on the matching result.
Notably, our methods can better handle the challenges such as textureless areas, repetitive patterns, and scale variations.


\subsection{Visual Localization}\label{4.4}
\textbf{Dataset and Metric.}
In this experiment, InLoc~\cite{taira2018inloc} and Aachen Day-Night v1.1~\cite{zhang2021reference} are used to verify the ability of our ASTR in visual localization.
InLoc~\cite{taira2018inloc} is an indoor dataset with 9972 RGBD images, of which 329 RGB images are employed as queries for visual localization.
The challenge of InLoc~\cite{taira2018inloc} mainly comes from textureless regions and repetitive patterns under large viewpoint changes.
In Aachen Day-Night v1.1~\cite{zhang2021reference}, 824 day-time images and 191 night-time images are chosen as queries for outdoor visual localization.
Large illumination and viewpoint changes pose challenges for Aachen~\cite{zhang2021reference}.
For both benchmarks, we evaluate the performance of our ASTR trained on MegaDepth~\cite{li2018megadepth} in the same way as ~\cite{sun2021loftr}.
The metrics of Inloc~\cite{taira2018inloc} and Aachen~\cite{zhang2021reference} are the same, which measure the percentage of images registered within given error thresholds.

\begin{table}[t]
	\centering
	\small
	\caption{Visual localization evaluation on the InLoc~\cite{taira2018inloc} benchmark.}
	\label{tab:Inloc_result}
	\vspace{-3mm}
	\scalebox{0.75}{
		\begin{tabular}{l c c}
			\hline
            \multicolumn{1}{c}{\multirow{2}{*}{Method}} & DUC1 & DUC2 \\
			\cline{2-3}
                     & \multicolumn{2}{c}{$\left(0.25m, 10^\circ\right)$ / $\left(0.5m, 10^\circ\right)$ / $\left(1m, 10^\circ\right)$} \\
			\hline
            Patch2Pix~\cite{zhou2021patch2pix}(w.SP~\cite{sarlin2020superglue}+CAPS~\cite{wang2020learning}) & 42.4 / 62.6 / 76.3 & 43.5 / 61.1 / 71.0 \\
            LoFTR~\cite{sun2021loftr} & 47.5 / 72.2 / 84.8 & 54.2 / 74.8 / \textbf{85.5} \\
            MatchFormer~\cite{wang2022matchformer} & 46.5 / 73.2 / 85.9 & \textbf{55.7} / 71.8 / 81.7 \\
            ASpanFormer~\cite{chen2022aspanformer} & 51.5 / \textbf{73.7} / 86.4 & 55.0 / 74.0 / 81.7 \\
            \textbf{ASTR(ours)} & \textbf{53.0} / \textbf{73.7} / \textbf{87.4} & 52.7 / \textbf{76.3} / 84.0 \\
			\hline
		\end{tabular}
	}
\end{table}

\begin{table}[t]
	\centering
	\small
	\caption{Visual localization evaluation on the Aachen Day-Night benchmark v1.1~\cite{zhang2021reference}.}
	\label{tab:Aachen_result}
	\vspace{-3mm}
	\scalebox{0.75}{
		\begin{tabular}{ l c c}
			\hline
            \multicolumn{1}{c}{\multirow{2}{*}{Method}} & Day & Night \\
			\cline{2-3}
                     & \multicolumn{2}{c}{$\left(0.25m, 2^\circ\right)$ / $\left(0.5m, 5^\circ\right)$ / $\left(1m, 10^\circ\right)$} \\
			\hline
			\multicolumn{3}{l}{\textbf{Localization with matching pairs provided in dataset}} \\
			\hline
            R2D2~\cite{r2d2}+NN & - & 71.2 / 86.9 / 98.9 \\
            ASLFeat~\cite{luo2020aslfeat}+NN & - & 72.3 / 86.4 / 97.9 \\
            SP~\cite{detone2018superpoint}+SuperGlue~\cite{sarlin2020superglue} & - & 73.3 / 88.0 / 98.4 \\
            SP~\cite{detone2018superpoint}+SGMNet~\cite{chen2021learning} & - & 72.3 / 85.3 / 97.9 \\
			\hline
			\multicolumn{3}{l}{\textbf{Localization with matching pairs generated by HLoc}} \\
			\hline
            LoFTR~\cite{sun2021loftr} & 88.7 / 95.6 / 99.0 & 78.5 / 90.6 / 99.0 \\
            ASpanFormer~\cite{chen2022aspanformer} & 89.4 / 95.6 / 99.0 & 77.5 / 91.6 / 99.0 \\
            AdaMatcher~\cite{huang2022adaptive} & 89.2 / \textbf{95.9}  / \textbf{99.2} & \textbf{79.1} / \textbf{92.1} / \textbf{99.5} \\
			\textbf{ASTR(ours)} & \textbf{89.9} / 95.6 / \textbf{99.2} & 76.4 / \textbf{92.1} / \textbf{99.5}    \\
			\hline
		\end{tabular}
	}
        \vspace{-3mm}
\end{table}

\textbf{Results.}
For InLoc~\cite{taira2018inloc} benchmark, our method achieves the best performance on DUC1 and is on par with state-of-the-art methods on DUC2 (in Tabel~\ref{tab:Inloc_result}).
For Aachen~\cite{zhang2021reference} benchmark, our ASTR performs comparative with others on Day and Night scenes (in Tabel~\ref{tab:Aachen_result}).
Overall, our method exhibits strong generalization ability in visual localization.

\subsection{Ablation Study}\label{4.5}

\begin{table}[t]
	\centering
	\small
	\caption{Ablation Study of each component on MegaDepth~\cite{li2018megadepth}.}
	\label{tab:ablation_study}
	\vspace{-3mm}
	\scalebox{0.75}{
		\begin{tabular}{c c c c c c c}
    		\hline
			\multirow{2}{*}{Index} & \multirow{2}{*}{Multi-Level} & \multirow{1}{*}{Spot-Guided} & \multirow{2}{*}{Scaling}
		    &\multicolumn{3}{c}{Pose estimation AUC} \\
			\cline{5-7}
                & &($l=5,k=4$) & & {@$5^\circ$}  &{@$10^\circ$}  &{@$20^\circ$} \\
		    \hline
		      1 & & & & 45.6 & 62.2 & 75.3 \\
		      2 & \checkmark & & & 46.7 & 63.1 & 76.3 \\
		      3 & \checkmark & \checkmark & & 47.7 & 64.5 & 77.4\\
		      4 & \checkmark & \checkmark & \checkmark & \bf 48.3 & \bf 65.0 & \bf 77.7\\
		     \hline
		\end{tabular}
	}
\end{table}

To deeply analyze the proposed method, we perform detailed ablation studies on MegaDepth~\cite{li2018megadepth} to evaluate the effectiveness of each component in ASTR.
Here, we use images with a size of 544 for training and evaluation.
As shown in Table~\ref{tab:ablation_study}, we intend to gradually add these components to the baseline.
The baseline (Index-1) we used is slightly different from LoFTR~\cite{sun2021loftr}.
More details can be found in Supplementary Material.

\begin{figure}[t]
    \centering
    \includegraphics[width=1.0\linewidth]{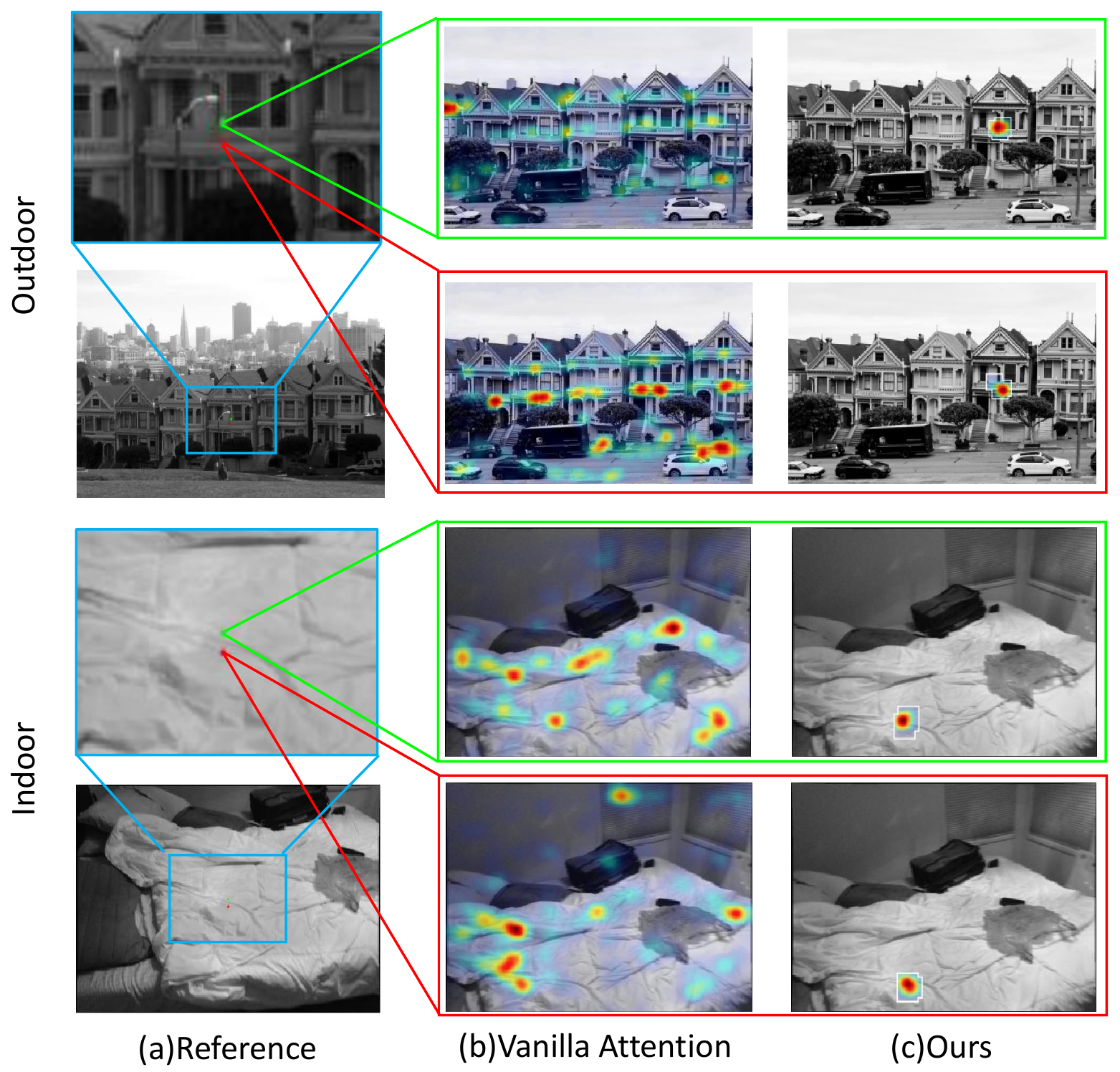}
    \vspace{-6mm}
	\caption{
	Visualization of vanilla and spot-guided cross attention maps on MegaDepth~\cite{li2018megadepth} (outdoor) and ScanNet~\cite{dai2017scannet} (indoor).
    }\label{fig:spot_vis}
\end{figure}

\textbf{Effectiveness of Spot-Guided Aggregation Module.}
We divide the spot-guided aggregation module into multi-level cross attention and spot-guided attention for ablation studies.
We first add vanilla cross attention layers at 1/32 resolution to the baseline (Index-2 in Table~\ref{tab:ablation_study}).
Comparing the results of Index-2 and Index-1, we conclude that 1/32 resolution global interaction across images is beneficial for image matching.
Then, in Index-3, linear attention layers at 1/8 resolution are substituted for the spot-guided attention layers.
The performance of Index-3 is improved compared with Index-2, which verifies the effectiveness of our spot-guided attention.
In Figure~\ref{fig:spot_vis}, we visualize vanilla and our spot-guided cross attention maps for contrast, showing that spot-guided attention can indeed avoid interference from unrelated areas.

\begin{table}[t]
	\centering
	\small
	\caption{Ablation Study with different $k$ and $l$ in spot-guided attention on MegaDepth~\cite{li2018megadepth}.}
	\label{tab:k_l_result}
	\vspace{-3mm}
	\scalebox{0.75}{
		\begin{tabular}{c c c c}
    		\hline
			\multirow{2}{*}{$k$($l=5$)} &\multicolumn{3}{c}{Pose estimation AUC} \\
			\cline{2-4}
                & {@$5^\circ$}  &{@$10^\circ$}  &{@$20^\circ$} \\
		    \hline
		      1  & 46.0 & 62.7 & 76.2 \\
			  2  & 47.5 & 64.0 & 77.1\\
		      3  & 47.3 & 63.8 & 76.7 \\
		      4  & \bf 47.7 & \bf 64.5 & \bf 77.4\\
			  5  & 47.1 & 63.7 & 77.0\\
			  6  & 46.9 & 63.6 & 76.6\\
		     \hline
		\end{tabular}
	}
	\scalebox{0.85}{
		\begin{tabular}{c c c c}
    		\hline
			\multirow{2}{*}{$l$($k=4$)} &\multicolumn{3}{c}{Pose estimation AUC} \\
			\cline{2-4}
                & {@$5^\circ$}  &{@$10^\circ$}  &{@$20^\circ$} \\
		    \hline
		      3  & 46.7 & 63.2 & 76.1 \\
			  5  & \bf 47.7 & \bf 64.5 & \bf 77.4\\
		      7  & 47.2 & 63.4 & 76.8 \\
		      9  & 43.0 & 60.5 & 74.8\\
		     \hline
		\end{tabular}
	}
\end{table}


To maximize the effectiveness of our spot-guided attention, we explore how to set suitable parameters $l$ and $k$.
First, in the setting of Index-3, we fix $l=5$ and vary $k$ from 1 to 6.
After observing the results in Table~\ref{tab:k_l_result}, the performance drops when $k$ is smaller than 4 or larger than 4.
Then, we fix $k=4$ and vary $l$ from 3 to 9.
As shown in Table~\ref{tab:k_l_result}, we find that the model achieves the best performance at $l=5$.
The reason may be that the spot area is too small to provide sufficient information from another image when using small $k$ or $l$.
With large $k$ or $l$, for a certain pixel, some matching areas of low confidence or dissimilar points will damage its feature aggregation.

\begin{figure}[t]
    \centering
    \includegraphics[width=1.0\linewidth]{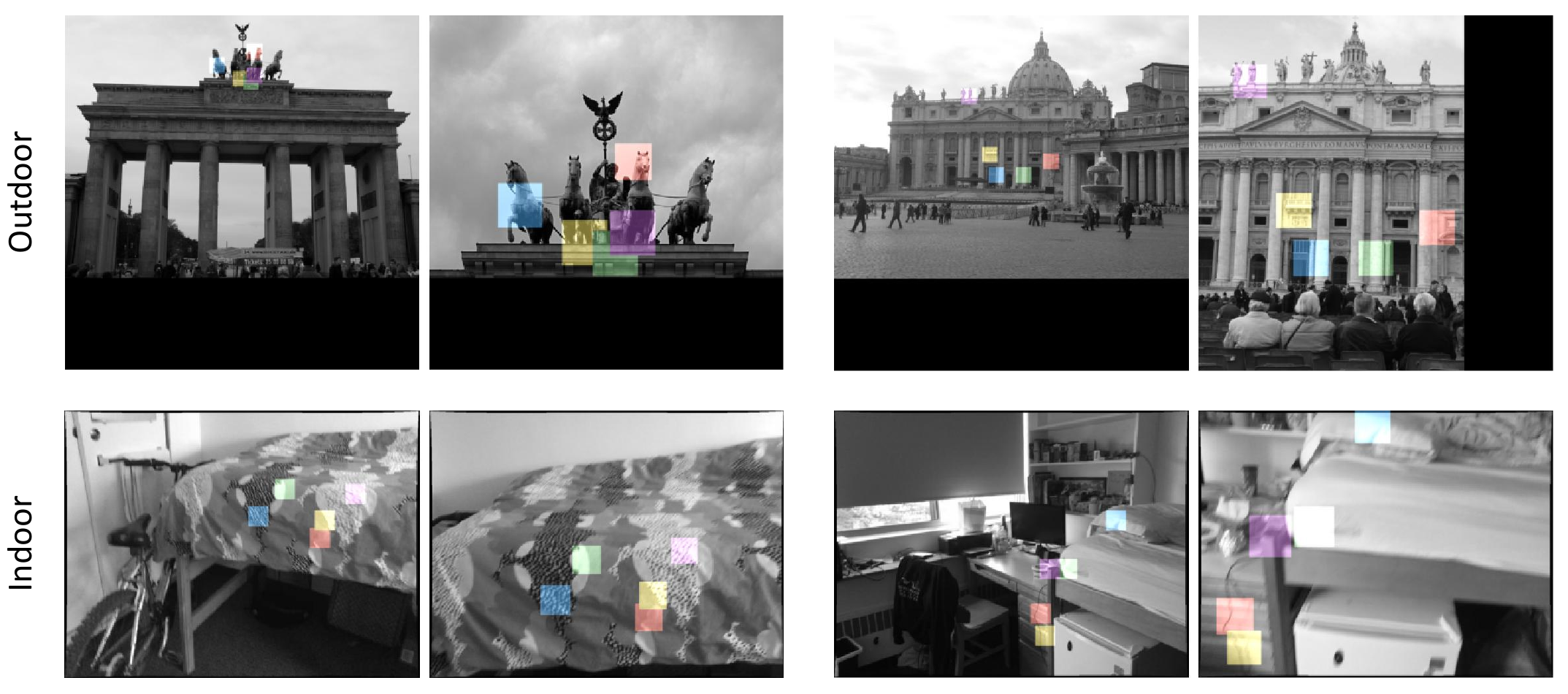}
    \vspace{-6mm}
	\caption{
	Visualization of grids from adaptive scaling module on MegaDepth~\cite{li2018megadepth} (outdoor) and ScanNet~\cite{dai2017scannet} (indoor).
    }\label{fig:grids_vis}
\end{figure}

\textbf{Effectiveness of Adaptive Scaling Module.}
As shown in Table~\ref{tab:ablation_study}, comparing the results of Index-4 and Index-3, we can see that the performance is improved, which indicates that coarse-level matching results are better refined with adaptive scaling module.
In Figure~\ref{fig:grids_vis}, we visualize the cropped grids from adaptive scaling module, indicating that our adaptive scaling module can adaptively crop grids of different sizes according to scale variations.

\section{Conclusion}
\label{sec:conclusion}
In this paper, we propose a novel Adaptive Spot-guided Transformer (ASTR) for consistent local feature matching.
To model local matching consistency, we design a spot-guided aggregation module to make most pixels avoid the impact of irrelevant areas, such as noisy and repetitive regions.
To better handle large scale variation, we use the calculated depth information to adaptively adjust the size of grids at the fine stage.
Extensive experimental results on five benchmarks demonstrate the effectiveness of the proposed method.

\noindent\textbf{Limitation.}
Although our adaptive scaling module is lightweight and pluggable, it demands camera pose estimation in the coarse stage, which requires the camera intrinsic parameters.
While camera intrinsic parameters are obtainable in standard datasets and most real-world scenarios, there are still some images from wild that lack them, rendering the adaptive scaling module disabled in such cases.


\clearpage

\twocolumn[{
\begin{center}
\textbf{
\Large Adaptive Spot-Guided Transformer for Consistent Local Feature Matching\\
\rule[3pt]{1.0cm}{0.1em}Supplementary Material\rule[3pt]{1.0cm}{0.1em}
}
\end{center}
}]

In this supplementary material, we first introduce the general sparse attention operator in Section~\ref{sec:1}.
In Section~\ref{sec:2}, we provide some details about our experiment.
In Section~\ref{sec:3}, we show additional visualizations about the spot-guided attention and adaptive scaling modules.

\begin{figure}[ht]
  \centering
  \includegraphics[width=1.0\linewidth]{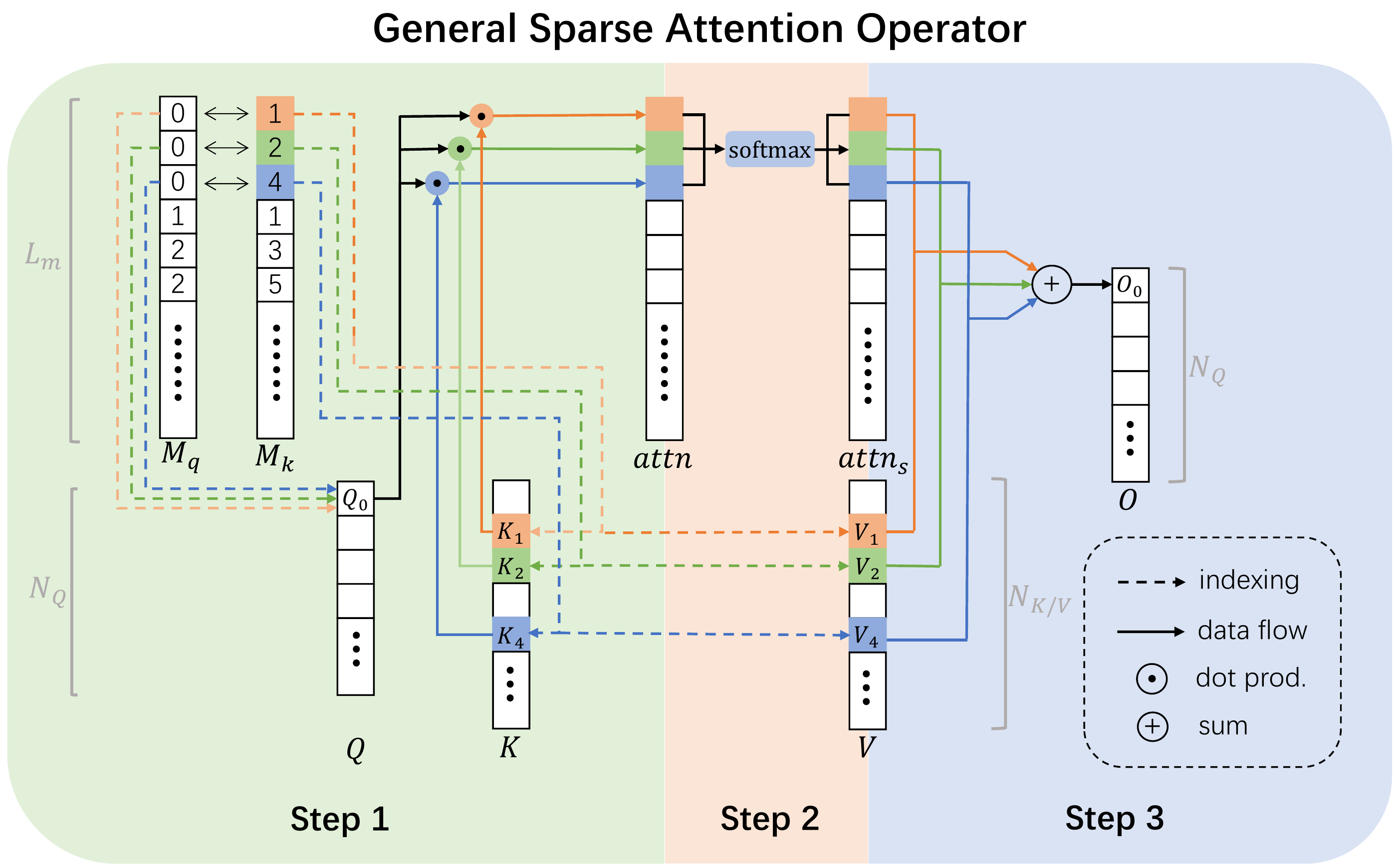}
  \vspace{-6mm}
\caption{
  The illustration of our general sparse attention operator.
  }\label{fig:sparse}
\end{figure}

\section{General Sparse Attention Operator}
\label{sec:1}

Due to irregular key/value token number for each query in Spot Attention, the naive implementation by PyTorch~\cite{paszke2019pytorch} is not efficient for memory and computation, which uses a mask to set unwanted values in the attention map to $0$.
More generally, the same problem also exists when the numbers of key corresponding to queries are not the same.
Inspired by PointNet~\cite{qi2017pointnet} and Stratified Transformer~\cite{lai2022stratified},
we implement a general sparse attention operator 
using CUDA that is efficient in terms of memory and computation.
We attempt to only compute the necessary attention between much less query/key tokens.

We can divide a vanilla attention operator into 3 steps.
Inputs are grouped as query $Q$, key $K$ and value $V$.
First, the attention map $A$ is computed by dot production as $A=QK^T$.
Then, a softmax operator is performed on the attention map: $A_s=\mathrm{softmax}(A/\sqrt{d_k})$.
Finally, the updated query $O$ can be obtained by $O=A_s V$.
We optimize  these three steps separately.

In the step 1, because only a few results in $A$ are useful for sparse attention, we do not need to compute the full $A$.
Instead, we compute the dot productions between $L_m$ pairs of query and key.
$M_q$ and $M_k$ record the indexes of query and key tokens whose dot productions are needed.
The length of $M_q$ and $M_k$ are both $L_m$.
Here, we denote the sparse attention map as $attn$, which is calculated by
\begin{equation}\label{eq:ep1}
\noindent attn[i]=Q[M_q[i]]K[M_k[i]]^T, \; i=0,1,\cdots, L_m-1.
\end{equation}

In the step 2, we group the elements in $attn$ with the same query index and apply $\mathrm{softmax}$ on each group.
The result is denoted as $attn_s$.

In the step 3, we compute the updated query
\begin{equation}\label{eq:ep2}
O[q]=\sum_{M_q[i]=q}{{attn}_s[i] \cdot V[M_k[i]]}.
\end{equation}

All of three steps are implemented in CUDA.

Compared with the naive implementation using PyTorch~\cite{paszke2019pytorch}, our highly optimized implementation reduces the memory and time complexity from $\mathcal{O}(N_q \cdot N_k \cdot N_h \cdot N_d^2)$ to $\mathcal{O}(L_m \cdot N_h \cdot N_d^2)$, where $N_q$, $N_k$ and $N_h$ are separately the numbers of query tokens, key tokens and attention heads, and $N_d$ is the dimension of each head.
Considering $L_m \ll N_q \cdot N_k$, our implementation is much more efficient than the naive implementation.

In particular, we also calculate the matching matrix in spot-guided attention in this way and set the probability of unrelated pixels to 0, which can greatly reduce the memory and computation cost.

\begin{figure}[t]
  \centering
  \includegraphics[width=0.75\linewidth]{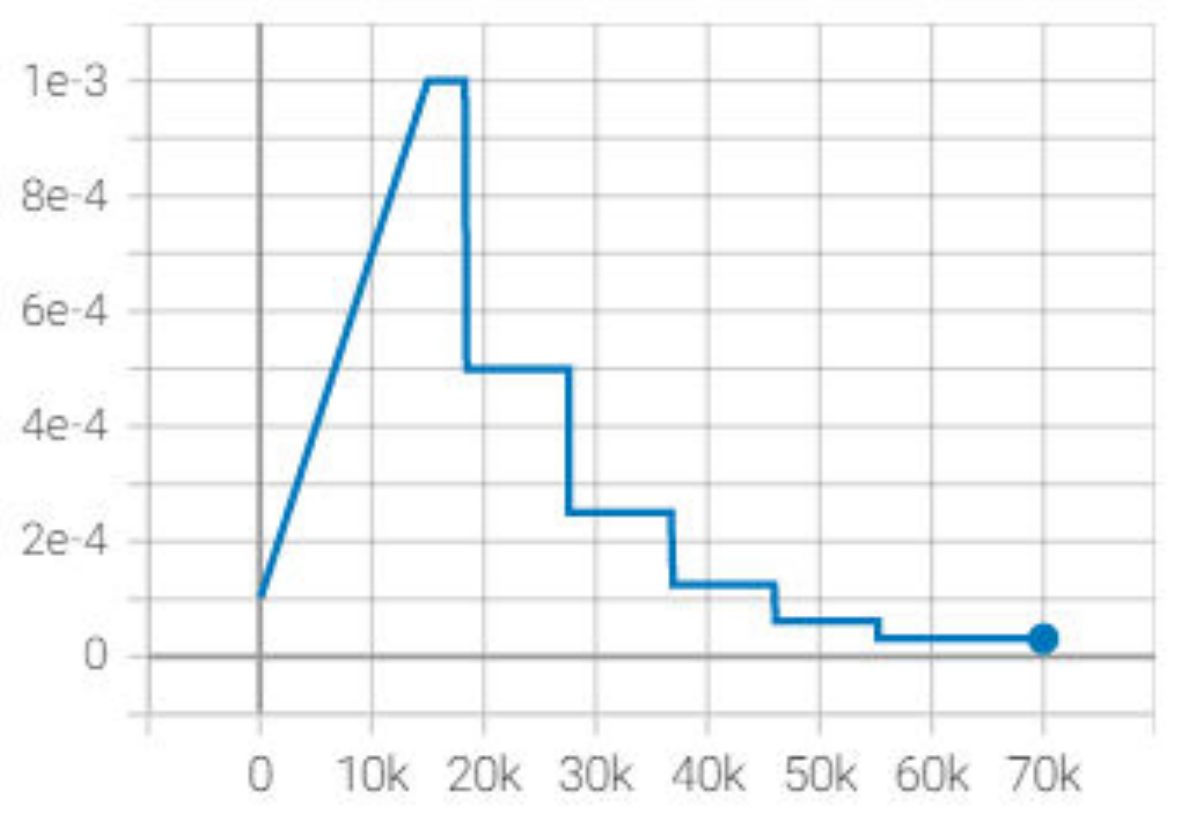}
  \vspace{-3mm}
\caption{
Learning rate curve while training on MegaDepth~\cite{li2018megadepth}.
  }\label{fig:learning_rate}
\end{figure}

\begin{figure}[t]
  \centering
  \includegraphics[width=1.0\linewidth]{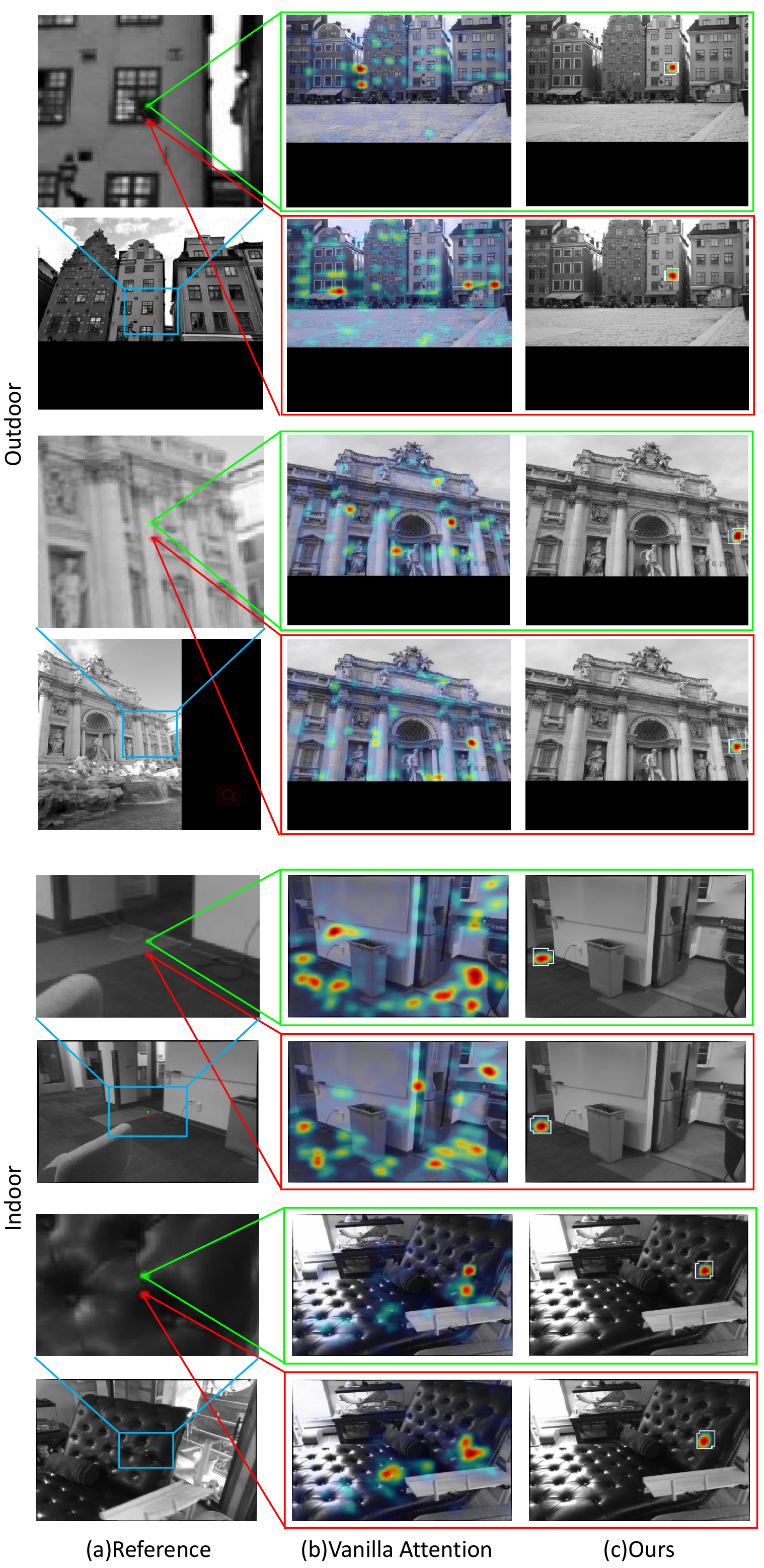}
  \vspace{-6mm}
\caption{
Visualizations of vanilla and spot-guided attention maps on MegaDepth~\cite{li2018megadepth} (outdoor) and ScanNet~\cite{dai2017scannet} (indoor).
  }\label{fig:spot}
\end{figure}

\section{Experimental Details}
\label{sec:2}

\subsection{Training Details}\label{sec:2.1}
To reduce the GPU memory, we randomly sample $50\%$ of ground truth matches to supervise the matching matrix at the coarse stage.
And we sample $20\%$ of the maximum number of coarse-level possible matches at the fine stage.
We train ASTR on MegaDepth~\cite{li2018megadepth} for 15 epochs.
The initial learning rate is $1 \times 10^{-3}$, with a linear learning rate warm-up for 15000 iterations.
The learning rate curve is shown in Figure~\ref{fig:learning_rate}.

\begin{figure}[t]
  \centering
  \includegraphics[width=1.0\linewidth]{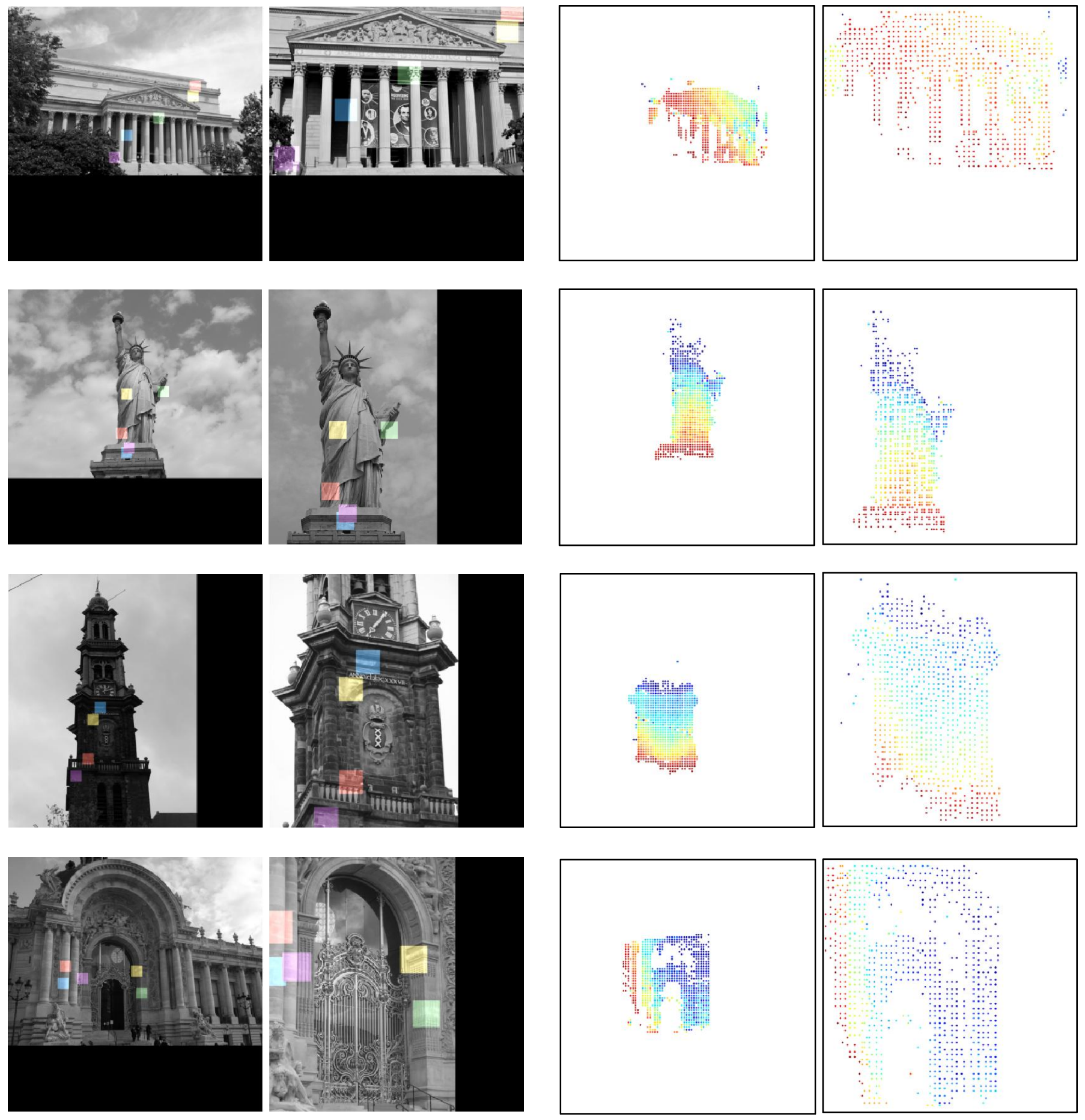}
  \vspace{-6mm}
\caption{
Visualizations of grids from adaptive scaling module and corresponding depth maps on MegaDepth~\cite{li2018megadepth}.
Note that we use depth values with scale uncertainty to compose the depth maps.
  }\label{fig:scale}
\end{figure}

\subsection{Differences between Baseline and LoFTR}\label{sec:2.2}
There are two main differences between our baseline and LoFTR~\cite{sun2021loftr}.

\textbf{(1) Normalized Positional Encoding.}
LoFTR~\cite{sun2021loftr} adopts the absolute sinusoidal positional encoding by following~\cite{carion2020end}:
\begin{equation}\label{eq:ep2}
\mathrm{PE}_i(x,y) = \left\{
  \begin{array}{ll}
  \mathrm{sin}(w_k \cdot x), & i = 4k \\
  \mathrm{cos}(w_k \cdot x), & i = 4k + 1 \\
  \mathrm{sin}(w_k \cdot y), & i = 4k + 2 \\
  \mathrm{cos}(w_k \cdot y), & i = 4k + 3 \\
  \end{array},
\right.
\end{equation}
where $w_k = \frac{1}{10000^{2k/d}}$, $d$ denotes the number of feature channels and $i$ is the index for feature channels.
Considering the gap in image resolution between training and testing, we utilize the normalized positional encoding as~\cite{chen2022aspanformer}, which is proven to mitigate the impact of image resolution changes in~\cite{chen2022aspanformer}.
The normalized positional encoding $\mathrm{NPE}_i(\cdot,\cdot)$ can be expressed as
\begin{equation}\label{eq:ep2}
\mathrm{NPE}_i(x,y) = \mathrm{PE}_i(x * \frac{W_{train}}{W_{test}}, y * \frac{H_{train}}{H_{test}}),
\end{equation}
where $W_{train/test}$ and $H_{train/test}$ are width and height of training/testing images.

\textbf{(2) Convolution in Attention.}
Chen et al.~\cite{chen2022aspanformer} find that replacing the self attention with convolution can improve the performance.
Hence, we deprecate self attention and MLP,  and utilize a $3 \times 3$ convolution in our ASTR.

\subsection{CNN Backbone}\label{sec:2.3}
Here we leverage a deepened version of Feature Pyramid Network (FPN)~\cite{lin2017feature}, which achieves a minimum resolution of 1/32.
The initial dimension for the stem is still 128 as LoFTR~\cite{sun2021loftr}, and the number of feature channels for subsequent stages is [128, 196, 256, 256, 256].

\section{Visualization Results}\label{sec:3}

In Figure~\ref{fig:spot}, we pick up two similar adjacent pixels as queries and visualize the corresponding attention maps of vanilla and our spot-guided attention for comparison.
The vanilla attention mechanism is vulnerable to repetitive textures, while our spot-guided attention can focus on the correct areas in these repeated texture regions.
Because large scale variation occurs frequently on outdoor datasets,
we mainly visualize the grids from the adaptive scaling module and corresponding depth maps on MegaDepth~\cite{li2018megadepth}.
As shown in Figure~\ref{fig:scale}, our adaptive scaling module can adjust the size of grids according to depth information.

{\small
\bibliographystyle{ieee_fullname}
\bibliography{CVPR2023bib}
}

\end{document}